\documentclass[11pt,a4paper]{article}
\usepackage[hyperref]{ranlp2023}
\usepackage{times}
\usepackage{latexsym}

\usepackage{microtype}
\usepackage{graphicx, subfigure}
\usepackage{algorithm}
\usepackage{algpseudocode}
\usepackage{makecell, multirow, multicol, amsmath}
\usepackage{tcolorbox}

\aclfinalcopy 

\title{Exposing Pink Slime Journalism: Linguistic Signatures and Robust Detection Against LLM-Generated Threats}

\author{
\textbf{Sadat Shahriar}, \textbf{Navid Ayoobi}, \textbf{Arjun Mukherjee}, \textbf{Mostafa Musharrat}, \textbf{Sai Vishnu Vamsi} \\
University of Houston, Texas, USA \\
\texttt{sadat.shrr@gmail.com}, \texttt{nayoobi@cougarnet.uh.edu}, \\\texttt{amukher6@central.uh.edu},\\ \texttt{\{mushsharat, saivishnuvamsis07\}@gmail.com}}

\date{}

\begin{document}
\maketitle
\begin{abstract}
The local news landscape, a vital source of reliable information for 28 million Americans, faces a growing threat from Pink Slime Journalism, a low-quality, auto-generated articles that mimic legitimate local reporting. Detecting these deceptive articles requires a fine-grained analysis of their linguistic, stylistic, and lexical characteristics. In this work, we conduct a comprehensive study to uncover the distinguishing patterns of Pink Slime content and propose detection strategies based on these insights. Beyond traditional generation methods, we highlight a new adversarial vector: modifications through large language models (LLMs). Our findings reveal that even consumer-accessible LLMs can significantly undermine existing detection systems, reducing their performance by up to 40\% in F1-score. To counter this threat, we introduce a robust learning framework specifically designed to resist LLM-based adversarial attacks and adapt to the evolving landscape of automated pink slime journalism, and showed and improvement by up to 27\%. 
\end{abstract}

\section{Introduction}

With local news outlets closing in massive numbers due to budget cuts \cite{rashidian2019friend}, it is creating a vacuum increasingly filled by automatically generated, low-quality, template-driven content--an insidious phenomenon, commonly referred to as ``Pink Slime'' (PS) journalism. Cloaked in the disguise of legitimacy, such local reporting is more often part of sprawling, centrally orchestrated networks, often designed to spread partisan agendas \cite{horne2024nela}. While prior research has examined their production \cite{horne2024nela} and social media amplification \cite{aljebreen2024analysis}, the accessibility of Large Language Models (LLM) and readily available technology enables anyone to launch new outlets and propagate PS articles, making the analysis of their linguistic patterns even more critical. 

\begin{figure}[t]
\centering
\includegraphics[width=.4\textwidth]{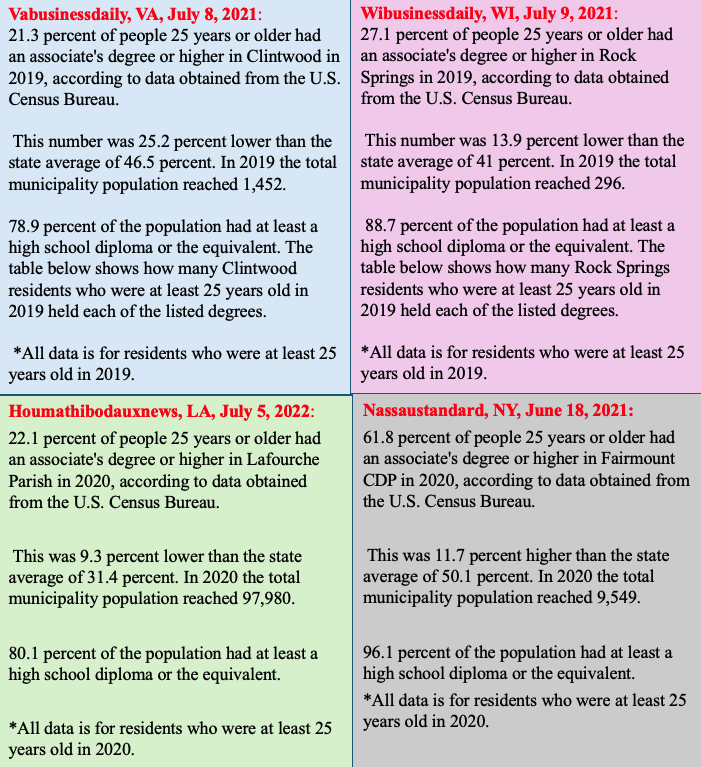}
\caption{Identical article templates are observed across four different news outlets, each targeting a different U.S. state and date. Only the location-specific statistics are varied. This formulaic, template-based approach is a hallmark of PS journalism, where local-seeming content is mass produced with minimal editorial variation.} 
\label{fig:PS_templates}
\end{figure}

First observed in the early 2010s, the term ``pink slime journalism'' was coined to describe the outlets that mimic legitimate local news (LN) while concealing their funding sources and ulterior motives \cite{carroll2019we,lepird2024pink,aljebreen2024analysis}. Furthermore, as shown in figure \ref{fig:PS_templates}, PS  articles frequently reuse identical templates with only superficial changes to names, locations, and statistics, revealing a low-effort, mass-produced approach to content generation.
Designed to mislead, manipulate public opinion, and serve hidden agendas, these articles pose a significant threat to the credibility and integrity of news ecosystems. 

To combat PS, researchers explored the tactics and strategies of the outlets \cite{horne2024nela}, and tweets and Facebook pages featuring them \cite{aljebreen2024analysis,lepird2023automated}. However, a research gap persists in leveraging the intrinsic textual properties of these articles for robust and scalable detection methods. Therefore we pose our first research question as \textbf{RQ1}:\textit{What are the defining linguistic, stylistic, and structural characteristics that distinguish pink slime articles from legitimate local news, and how can these differences be leveraged for accurate detection?} In our study, we observe, features such as lexical richness, and syntactic structures highlight the simplicity and repetitiveness often found in PS articles, while Parts-of-speech (POS) and dependency co-occurrence probabilities uncover their distinct grammatical patterns. 

While these textual features are effective against conventional PS articles, the rapid emergence of LLMs presents a new challenge. Studies show that journalists may feed confidential inputs, such as private correspondence or third-party articles, into LLMs to generate publishable content with little human oversight \cite{brighamdeveloping}. Moreover, LLMs can act as misinformation engines \cite{pan2023risk}, helping PS outlets mask their low-effort patterns and evade detection.

This leads to our second research question:
\textbf{RQ2}: \textit{Can detection models adapt to LLM-generated evasive pink slime content?}
To explore this, we simulate a targeted ``modification attack'' where LLMs rewrite PS articles to obscure their typical traits. Our results show detection performance drops by up to 40\%. To address this, we introduce a continual learning-based adaptation framework that incrementally retrains the model on both original and LLM-modified PS content. This approach restores performance, yielding a 27\% improvement in detecting adversarially rewritten PS articles.







Our main contributions are:

\begin{itemize}
    \item This is the first work that provides an in-depth textual analysis of Pink Slime news articles and explains what sets them apart from legitimate local news.

    \item We investigate the vulnerabilities of existing detection models against LLM-modified adversarial PS articles, and propose a continual learning-based framework that significantly enhances detection robustness.
\end{itemize}

\section{Related Works}
The imminent threat of PS journalism has been widely acknowledged \cite{moore2023consumption,carroll2019we,bengani2019hundreds}, yet only a limited number of studies have systematically addressed detection strategies. Lepird and Kathleen proposed an automated, network-based detection approach that identifies potential PS sites by analyzing their social media dissemination patterns \cite{lepird2023automated}. Similarly, Aljabreen et al. examined automated dissemination dynamics, uncovering distinct linguistic and structural features in how PS content is shared, such as truncated first sentences and minimal user-generated commentary \cite{aljebreen2024analysis}.

Horne and Gruppi introduced the NELA-PS dataset, which we utilize in this research. The dataset contains articles with metadata such as source, location, network, and IP addresses collected over 2.5 years \cite{horne2024nela}. Their study provides a detailed examination of PS article production, analyzing metrics such as the number of articles produced daily per source and cross-platform content sharing. They also performed production-based comparisons between legitimate local news and Pink Slime content, offering critical insights into the mechanics of Pink Slime journalism. These findings are invaluable for understanding Pink Slime production and devising strategies to combat it.

However, none of the aforementioned studies have conducted a comprehensive textual content analysis or developed detection strategies focused on the linguistic and stylistic properties of PS articles. This represents a significant and critical research gap in this domain.

\section{Datasets}



We use the Pink Slime dataset curated by Horne and Gruppi \cite{horne2024nela}, containing 7.9 million articles from 1,093 outlets (2021–2023). From this, we randomly sample 40,000 articles and apply deduplication using all-MiniLM-L6-v2 embeddings with a cosine similarity threshold of 0.8, reducing the size by 76.25\% due to repeated templates (e.g., crime, gas prices) and cross-network copying. After removing NaNs, we retain 9,473 PS articles. We similarly process local news articles from \cite{horne2022nela}, ending up with 10,000 legitimate samples, with only a 9.81\% reduction.

We also include the LIAR dataset \cite{wang-liar} with fake news annotations \cite{upadhayay2020sentimental}, and the NELA-GT-2021 dataset \cite{nelagt2021}, restricted to U.S. articles. In NELA-GT-2021, articles with factuality scores $\geq 4$ are labeled legitimate, and those with $\leq 1$ as fake news, providing diverse annotated content.

For experiments, we applied DBSCAN on the t-SNE representations (see Sec. \ref{sec:comparing}) using a minimum sample size of 10. Despite deduplication, many PS articles retained templated formats, so we avoided train-test overlap by selecting entire clusters—randomly chosen to form 80\% of the PS data for training, with the rest for testing. LN articles showed no clear clustering, so we applied a random 80-20 split. This process was repeated three times to ensure stability and reduce sampling bias.

\section{What Sets Apart Pink Slime from Legitimate Local News}


\subsection{Linguistic and Syntactic Simplicity in Pink Slime Articles}

\begin{figure*}[htbp]
    \centering
    \subfigure[]{\includegraphics[width=0.32\textwidth]{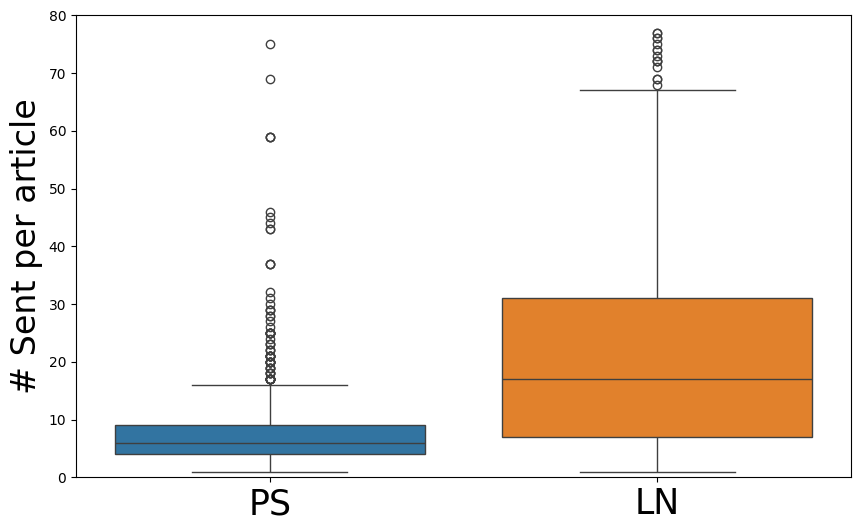}} 
    \subfigure[]{\includegraphics[width=0.32\textwidth]{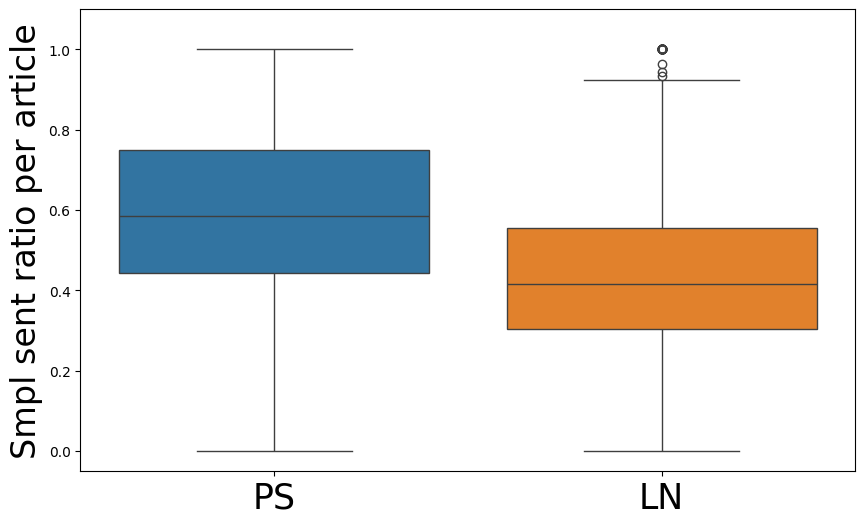}}
    \subfigure[]{\includegraphics[width=0.32\textwidth]{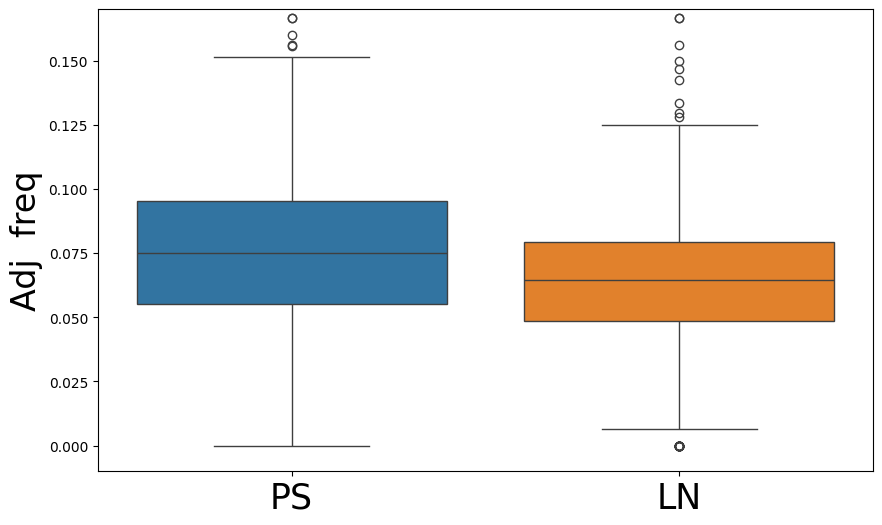}}
    \subfigure[]{\includegraphics[width=0.32\textwidth]{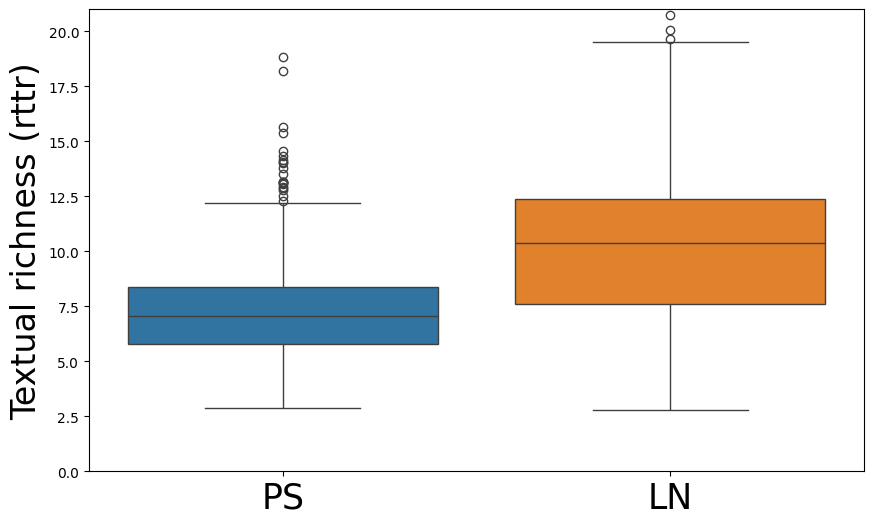}} 
    \subfigure[]{\includegraphics[width=0.32\textwidth]{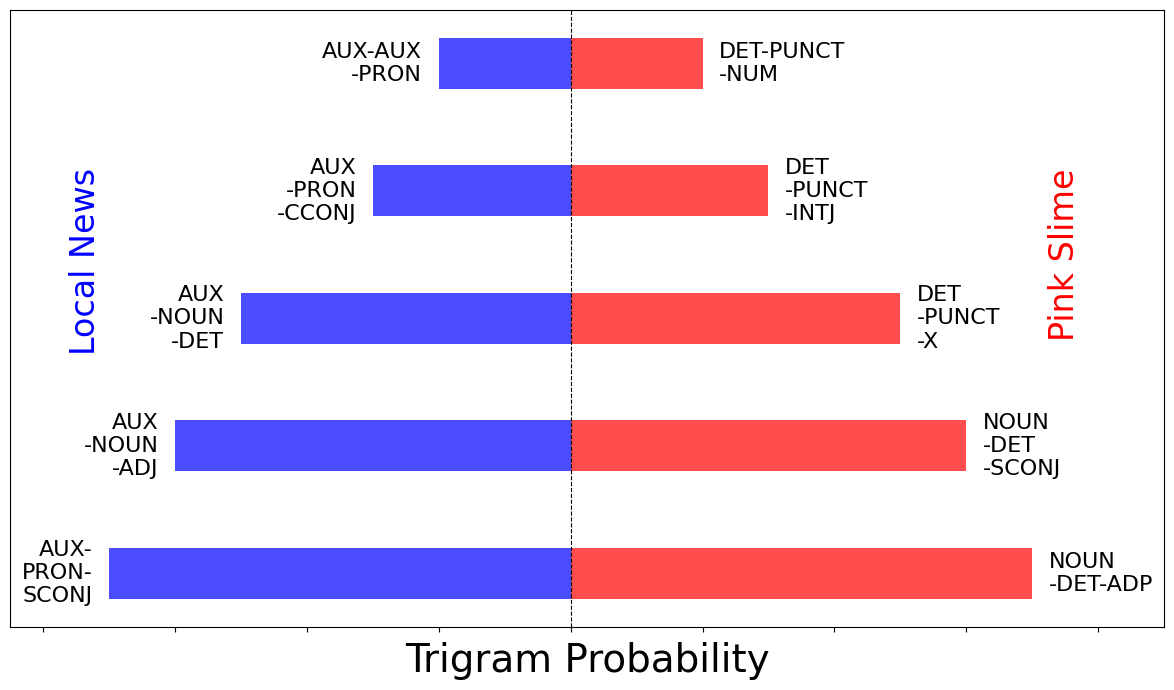}}
    \subfigure[]{\includegraphics[width=0.32\textwidth]{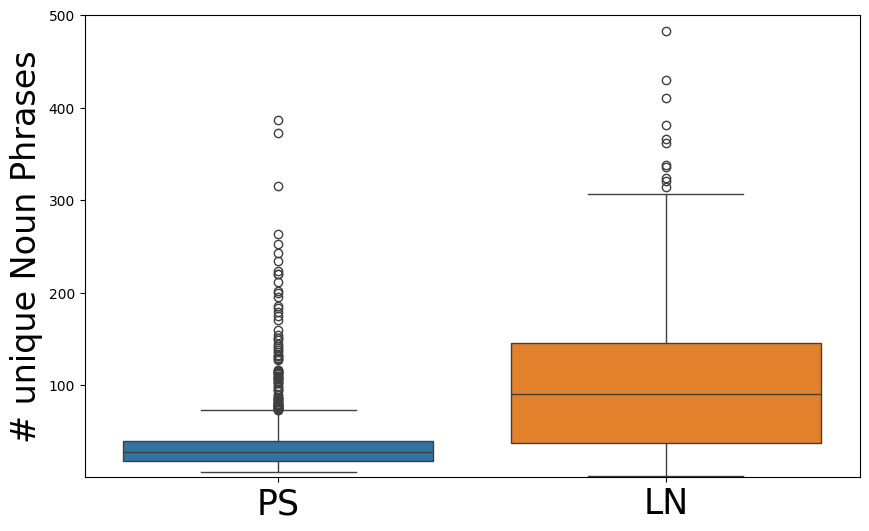}} 

    \caption{Comparison of key characteristics between regular local news (LN) and Pink Slime (PS): (a) Number of sentences per article, (b) Proportion of simple sentences, (c) Adjective frequency per 1000 words, (d) Lexical richness (Root Type-Token Ratio, RTTR), (e) Top 5 POS trigram probabilities, and (f) Number of unique noun phrases. }
    \label{fig:linguistic_analysis}

\end{figure*}

PS articles are significantly shorter than legitimate LN articles, averaging 8.8 vs. 22.97 sentences per article (p = 0.00; Fig. \ref{fig:linguistic_analysis}a). This brevity reflects PS’s cost-efficient, low-effort reporting style \cite{horne2024nela}. PS also uses more simple sentences (60.10\% vs. 44.34\%, p = 0.00; Fig. \ref{fig:linguistic_analysis}b) and more adjectives (75 vs. 65 per 1,000 words). Such simplicity aligns with PS’s strategy to reduce production effort, avoid complexity, and provide audiences with limited time. Additionally, PS articles employ more adjectives (75 per 1,000 words) than LN (65 per 1,000 words), suggesting an emphasis on descriptive, emotionally charged language to sensationalize content (Figure \ref{fig:linguistic_analysis}c).

Lexically, PS shows lower richness. We used the Root-Type-Token Ratio (RTTR) \cite{guiraud1954}, and found PS articles contain much less richness (RTTR = 7.26) than LN (10.16) and national news (10.21), suggesting limited vocabulary (Fig. \ref{fig:linguistic_analysis}d). Our analysis (Figure \ref{fig:linguistic_analysis}f) also demonstrates that PS articles have a significantly lower count of unique noun phrases compared to LN, further underscoring the reduced linguistic complexity and diversity in pink slime content.

Syntactically, LN frequently uses complex POS trigrams. As shown in figure \ref{fig:linguistic_analysis}e, many top trigrams in LN begin with AUX, followed by other tags like PRON (Pronoun), NOUN (Noun), or PROPN (Proper Noun). This suggests that legitimate local news may often use auxiliary verbs to form complex sentence structures, indicating a more structured and informative style. Examples include (`AUX', `PRON', `SCONJ') and (`AUX', `PRON', `CCONJ'), where auxiliary verbs are combined with pronouns and subordinating or conjunctions, hinting at complex sentences with dependent clauses for clarity or elaboration. In contrast, PS relies heavily on simple syntactic structures, frequently using noun phrases (NOUN, DET) followed by prepositions (ADP) or subordinating conjunctions (SCONJ). Additionally, the presence of punctuation-based trigrams (DET, PUNCT, X) highlights a less formal or fragmented writing style.

\begin{table*}[h!]
\caption{List of hand-crafted features and their descriptions.}
\small
\centering
\renewcommand{\arraystretch}{1.5}
\begin{tabular}{|p{5cm}|p{10cm}|}
\hline
\textbf{Feature Category} & \textbf{Description} \\ \hline
Word and sentence count & Count of total sentences and number of words per sentence in each news article \\ \hline
lexical richness & Lexical richness measured by Root Type-Token Ratio (RTTR),  corrected Type-Token Ratio (CTTR), and the measure of Measure of Textual Lexical Diversity (MTLD) \\ \hline

\# Simple sentence & Count of simple sentences per article. \\ \hline

POS co-occurrence probability & Probability of POS tags co-occurrence. we selected the top four POS co-occurrences with the greatest differences between PS and LN. These co-occurrences include space-verb, space-proper noun, unknown-space, and interjection-verb. \\ \hline

Dependency tag co-occurrence probability & Probability of dependency tags co-occurrence. We selected the top four dependency co-occurrences with the greatest differences between PS and LN. These co-occurrences include space-verb, space-proper noun, unknown-space, and interjection-verb. \\ \hline

Dependency parsing & Depth, average branching factor, and the length of the longest dependency chain of the syntactic dependency tree.\\ \hline

Readability & Flesch reading ease score. \\ \hline

POS count & Count of unique noun phrases, frequency of adjectives, and adverbs \\

 \hline
\end{tabular}
\label{table:hand_crafted_features}
\end{table*}

\subsection{Comparing Embedding-Based Clustering of News Article Types}
\label{sec:comparing}

\begin{figure}
\centering
\includegraphics[width=.4\textwidth]{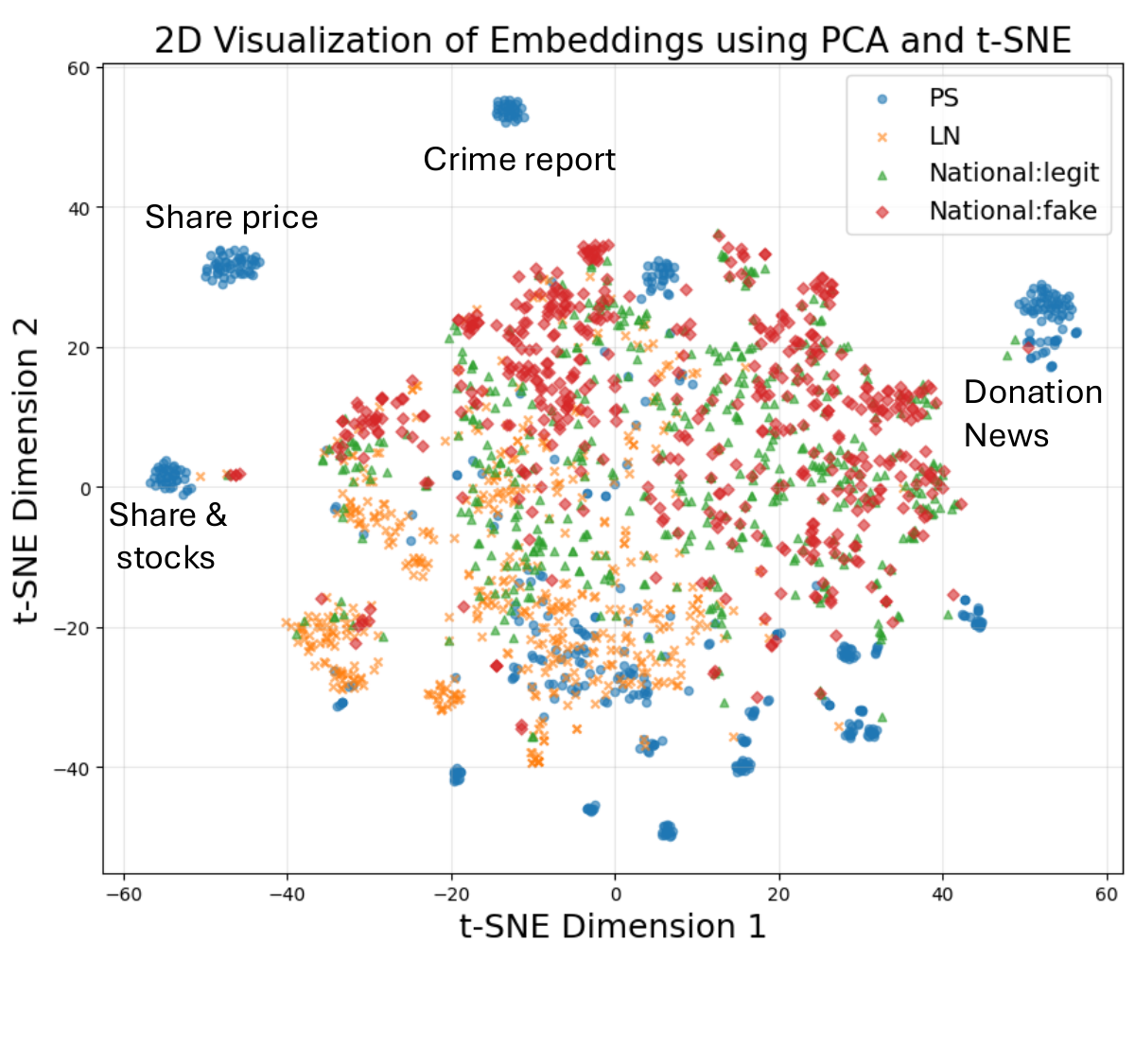}
\caption{t-SNE visualization of Pink Slime (PS), local news (LN), legitimate national news (National:legit), and fake news (National:fake). The dense clusters of PS are also indicated.} 
\label{fig:tsne}
\end{figure}

To compare PS articles with other news sources like LN, legitimate national news, and fake news--we utilize a textual embedding representation generated by the \textit{stella\_en\_400M\_v5} model \cite{zhang2024stella}. This model, with 435M parameters, ranks 6th on the MTEB leaderboard and is the top-performing model among those with fewer than 1.5B parameters \cite{muennighoff2022mteb}. To visualize the embeddings, we first reduce their dimensionality to 50 using Principal Component Analysis (PCA), followed by t-SNE to project them into a 2D space for better interpretability.

As shown in Figure \ref{fig:tsne}, PS articles exhibit a distinct pattern of dense micro-clusters scattered across the 2D representation. Upon further investigation, these clusters correspond to closely related articles covering topics such as stock performance, crime reports, donation news, and similar themes. The proximity and uniformity of these clusters in terms of linguistic structure suggest templated writing, as noted in prior studies \cite{horne2024nela}.

In contrast, LN articles form fewer but more dispersed clusters, reflecting greater diversity in content and style without the formation of tightly packed groups. Legitimate national news demonstrates an even broader distribution, indicating highly varied content and less structural uniformity. Interestingly, fake news articles deviate significantly from PS articles, with their embeddings also exhibiting a dispersed pattern similar to LN and national news, further emphasizing their divergence from the dense micro-clustered nature of PS articles.

\section{Detecting Pink Slime Articles}

\begin{figure}[htbp]
    \centering
    \subfigure[]{\includegraphics[width=0.40\textwidth]{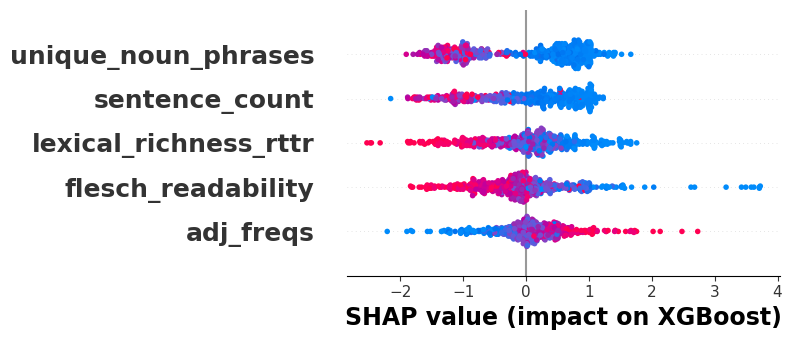}} 
    \subfigure[]{\includegraphics[width=0.40\textwidth]{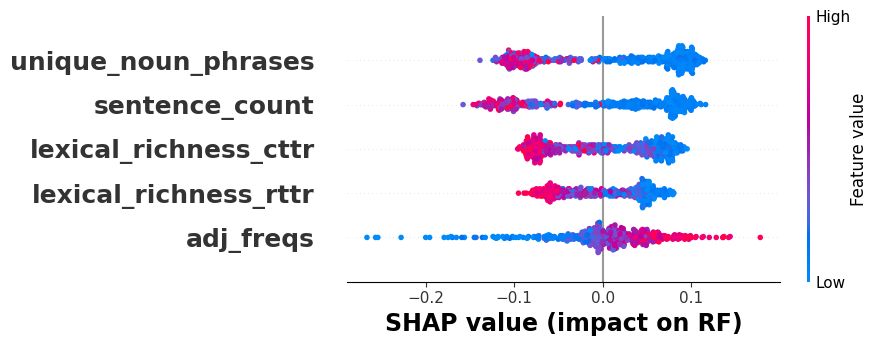}}

     \caption{ SHAP summary plots for the top features identified by the XGBoost (a) and Random Forest (b) classifiers. The x-axis shows the SHAP value (impact on model output), and features are ranked by their importance.}
    \label{fig:SHAP}

\end{figure}

We use different approaches to detect and differentiate PS from LN. First, we explore handcrafted features, focusing on structural and linguistic attributes rather than semantic content. Table \ref{table:hand_crafted_features} provides a detailed description of these features. For the classification task, we use three algorithms: XGBoost, Random Forest, and SVM, with Random Forest showing slightly better performance than the others. Figure \ref{fig:SHAP} compares SHAP summary plots for feature contributions in our task using XGBoost (a) and Random Forest (b). The x-axis shows SHAP values, indicating each feature's impact on the model's predictions, with positive values favoring PS and negative values favoring LN.

We found that both models demonstrate agreement in assigning high importance to features such as ``unique noun phrases'' and ``sentence count,'' suggesting these are key discriminative signals in the classification task. As discussed previously, higher counts of unique noun phrases likely indicate more diverse or structured content, which may be characteristic of legitimate LN articles, while simpler sentence structures (lower ``sentence count'') might be a hallmark of PS content. Differences arise in how the models treat readability metrics and lexical richness: XGBoost assigns a stronger influence to Flesch readability and RTTR, possibly leveraging its flexibility in capturing non-linear relationships, whereas Random Forest emphasizes CTTR, reflecting its sensitivity to slightly different patterns in feature splitting. These findings indicate the important role of linguistic diversity, syntactic complexity, and lexical structure in identifying PS articles. 

Next, we apply a transformer-based full fine-tuning approach using models like BERT \cite{devlin2019bert}, XLNet \cite{yang2019xlnet}, and Flan-T5 \cite{longpre2023flan} for the classification task. These models are fine-tuned on the full text of articles, allowing them to capture both semantic and contextual information, surpassing the limitations of handcrafted features that only focus on structural and linguistic cues. Among the models, BERT shows slightly better performance (89.31\% in F1-score) compared to XLNet and Flan-T5, although the difference is not statistically significant. The fine-tuning process updates all model weights during training, enabling the models to learn task-specific representations directly from the data. Additionally, a text embedding approach combined with a downstream fully connected (FC) model performed better than the handcrafted features but fell short of full fine-tuning. Both approaches benefit from leveraging semantic and syntactic characteristics of textual data, offering an edge since many PS articles reuse similar content across different sources. In contrast, handcrafted features rely solely on syntactic attributes, making it more challenging to detect PS content effectively.

\begin{table*}[h!]
\caption{Performance comparison for classifying Pink Slime (PS) vs regular Local News (LN). The hand-crafted features are mostly what is discussed in Figure 1. }
\small
\centering
\renewcommand{\arraystretch}{1.2}
\begin{tabular}{l|l|c|c}
\hline
\multicolumn{2}{c|}{\textbf{Model}} & \textbf{Accuracy (\%)} & \textbf{F1-score (\%)} \\ \hline
\multirow{3}{*}{\makecell{Hand-\\crafted\\ features}} & XGBoost & 78.00  & 77.46  \\ \cline{2-4} 

 & Random Forest & 78.57  & 79.86  \\ \cline{2-4} 
 & SVM & 74.38  & 74.01 \\ \hline
\multirow{3}{*}{fine-tuning} & BERT & 89.31  & 89.05    \\ \cline{2-4} 
& XLNet & 87.68  & 88.44    \\ \cline{2-4} 
& FlanT5 & 87.14   & 87.83  \\ \hline
\multirow{3}{*}{\makecell{embedding +FC}} & all-miniLM-L6-v2 & 82.32  & 83.56    \\ \cline{2-4} 
& gte-base-en-1.5 & 84.81  & 85.56   \\ \cline{2-4} 
& stella\_en\_400M\_v5 & 88.18  & 88.11    \\ \hline
\end{tabular}
\label{table:ps_vs_local}
\end{table*}

\section{Impact of LLM in pink slime journalism}

To simulate adversarial scenarios, we rephrased PS articles in ways that mimic strategies PS journalists might use to improve their quality and evade detection. Based on our observations of linguistic and structural patterns in PS articles, we applied LLM-based paraphrasing techniques aimed at enhancing journalistic quality. Specifically, our prompt strategies targeted expansion of content, incorporation of complex and compound sentence structures, reduction of excessive adjectival usage, enforcement of a neutral tone, introduction of diverse vocabulary, and increased frequency of unique noun phrases. Each criterion was intended to obfuscate detection mechanisms trained on surface-level cues.

We employed a mix of open-source (Gemma \cite{team2024gemma}, LlaMA \cite{touvron2023llama}, Qwen \cite{bai2023qwen}) and closed-source API-based models (GPT-4o-mini \cite{openai2023gpt4}, Claude-3.5-Haiku \cite{anthropic2024claude3}) to generate adversarial variants. To introduce more variety, we used Gemma 2B, Qwen and Mistral  (v0.3) and Llama-3.2 1B, 3B, and Llama3- 8B models. As shown in table \ref{tab:rephrasing_metrics}, among open-source models, Qwen was the most effective at evading detection: when tested against a previously fine-tuned BERT classifier (baseline F1-score: 89\%), it led to a 28.34\% drop in detection performance (Table \ref{tab:rephrasing_metrics}). Closed-source models were significantly more potent, with both GPT-4o-mini and Claude-3.5-Haiku reducing the F1-score by an average of 40\%, highlighting the challenge posed by high-quality LLM-generated obfuscation.

\begin{table}[htbp]
\caption{Pink slime detection accuracy if these articles are modified using different LLMs.}
\small
\centering
\begin{tabular}{cc}
\hline
LLM & F1-score \\
\hline
Gemma-2B & 68.57 \\
Mistral-7b & 65.63 \\
Llama-1b & 66.47 \\
Llama-3b & 62.76 \\
Llama-8b & 61.50 \\
Qwen-7b & 60.71 \\
\hline
Gpt-4o-mini & 48.15\\
Claude-3.5-haiku & 49.55\\
\hline
\end{tabular}
\label{tab:rephrasing_metrics}
\end{table}

\subsection{Model adaptation for LLM-attack}
\begin{figure*}
\centering
\includegraphics[width=.8\textwidth]{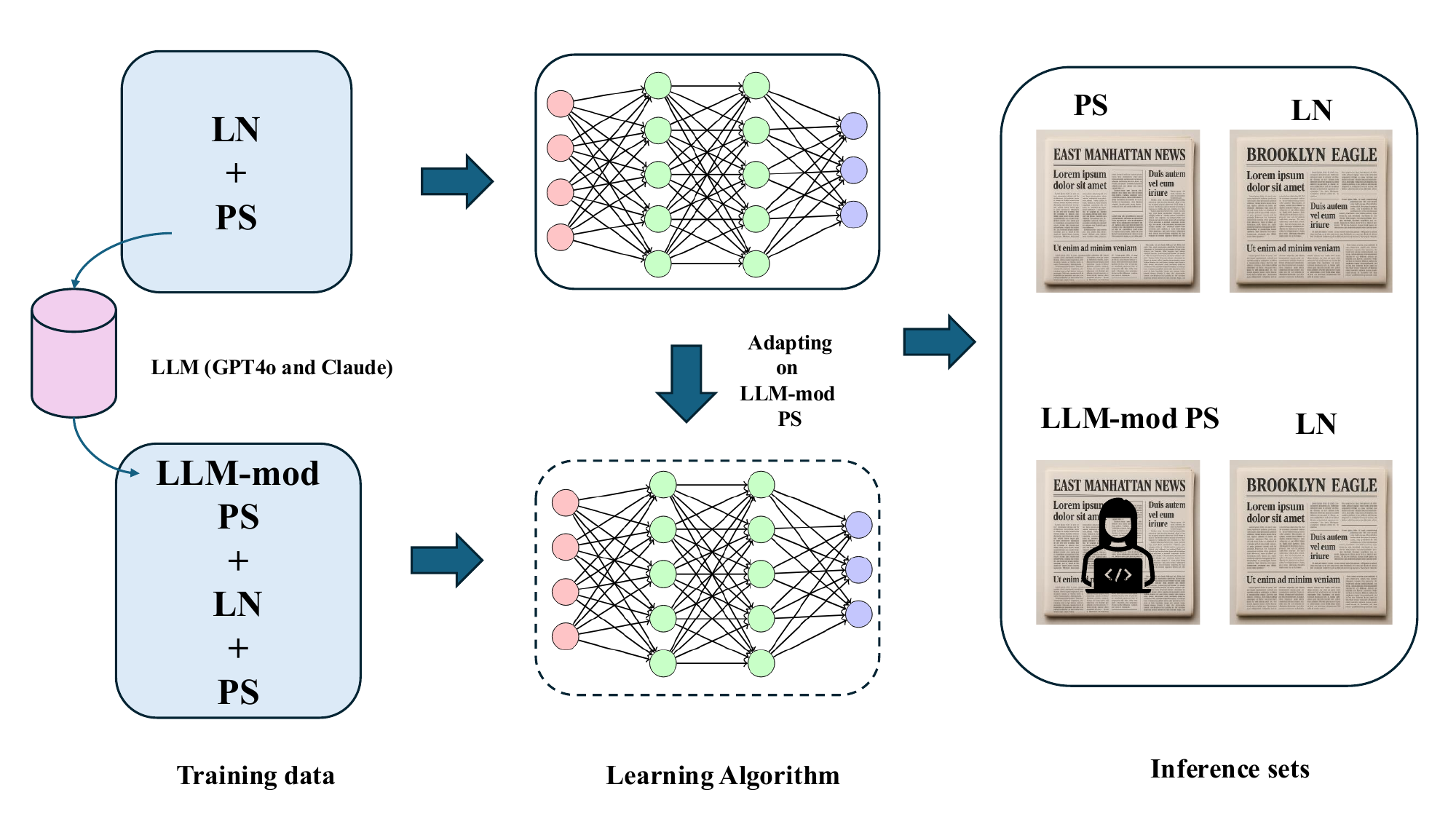}
\caption{First, the Model training involves a dataset combining LN and PS articles. To simulate adversarial scenarios, targeted LLM-paraphrased PS articles are introduced. Evaluation is performed on two test sets: the original set with human-written PS articles, and the LLM-modified set where only PS samples are paraphrased, while LN remains unchanged.} 
\label{fig:CL-flow}
\end{figure*}

To improve the detection performance, we employ a continual learning approach. We take the already trained model (we used BERT), and keep on training it with 1/100th learning rate with new data. We gradually train it with both original LN, and LLM-modified PS data. We take 10\% , 20\%  to 100\% of the data. To avoid forgetting, we also incorporate original PS data, which is 50\% of the PS sample. Figure \ref{fig:CL-flow} demonstrates the model training process. 

\subsection{Model Adaptation under LLM-based Adversarial Drift}

To enhance the robustness of the detection model against LLM-generated adversarial perturbations, we adopt a continual learning paradigm tailored for concept drift adaptation. Let $\mathcal{D}_{\text{base}} = \{(x_i, y_i)\}_{i=1}^{N}$ denote the original training dataset, where $x_i$ are input news samples and $y_i \in \{0,1\}$ denote labels corresponding to LN and PS, respectively. After initial fine-tuning of a pre-trained model $\theta_0$ on $\mathcal{D}_{\text{base}}$, we denote the trained parameters as $\theta^{(0)}$.

To adapt the model to adversarially modified PS samples produced via LLM paraphrasing (denoted as $\mathcal{D}_{\text{adv}}$), we employ \textit{incremental fine-tuning} on mixtures of $\mathcal{D}_{\text{adv}}$ and the original LN/PS distributions. Specifically, the training set at stage $t$ is:

\begin{equation}
\mathcal{D}^{(t)} = \mathcal{D}^{(t)}_{\text{LN}} \cup \mathcal{D}^{(t)}_{\text{adv}} \cup \mathcal{D}^{(t)}_{\text{PS-base}},
\end{equation}

where:
\begin{itemize}
    \item $\mathcal{D}^{(t)}_{\text{adv}} \subset \mathcal{D}_{\text{adv}}$ is an incrementally growing subset comprising $t\%$ of the total adversarial data,
    \item $\mathcal{D}^{(t)}_{\text{LN}}$ is the set of original LN, sampled proportionately, i.e.,  
    $|\mathcal{D}^{(t)}_{\text{LN}}| \approx |\mathcal{D}^{(t)}_{\text{adv}}| +  |\mathcal{D}^{(t)}_{\text{PS-base}}|$

    \item $\mathcal{D}^{(t)}_{\text{PS-base}}$ contains $50\%$ of the original PS samples, i.e., 
    $|\mathcal{D}^{(t)}_{\text{PS-base}}| = 0.5 \cdot |\mathcal{D}^{(t)}_{\text{adv}}|$, incorporated to mitigate catastrophic forgetting.
\end{itemize}

At each step $t$, we continue training with a reduced learning rate $\eta_t = \eta_0 / 100$, where $\eta_0$ is the original fine-tuning learning rate. Model parameters are updated via gradient descent:

\begin{equation}
\theta^{(t)} \leftarrow \theta^{(t-1)} - \eta_t \cdot \nabla_{\theta} \mathcal{L}(\theta^{(t-1)}; \mathcal{D}^{(t)}),
\end{equation}

where $\mathcal{L}$ denotes the cross-entropy loss computed over the composite dataset $\mathcal{D}^{(t)}$.

This progressive continual adaptation strategy allows the model to gradually incorporate distributional shifts induced by LLM-generated adversarial examples, and retain discriminative capacity on the original PS class through rehearsal on $\mathcal{D}^{(t)}_{\text{PS-base}}$. As such, it also stabilizes training via a reduced learning rate to prevent overfitting on small adversarial batches.

Our empirical results demonstrate that this approach significantly improves generalization under adversarial drift while maintaining accuracy on the original test distribution. For experimental purpose, we refer it as the \textit{controlled} setting. For comparison, we also define an \textit{uncontrolled} setting, where no rehearsal on original PS data is used, i.e., $\mathcal{D}^{(t)}_{\text{PS-base}} = \emptyset$. 
\begin{figure}[htbp]
    \centering
    \subfigure[]{\includegraphics[width=0.24\textwidth]{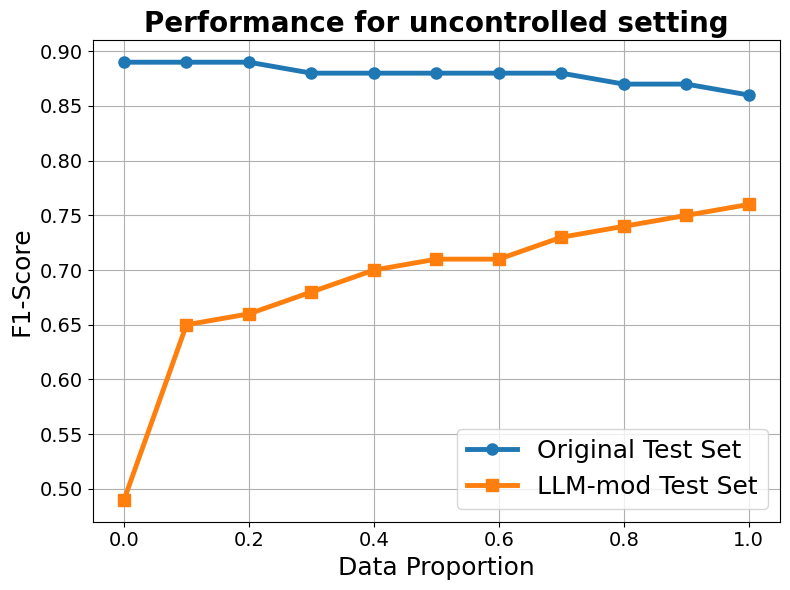}} 
    \subfigure[]{\includegraphics[width=0.23\textwidth]{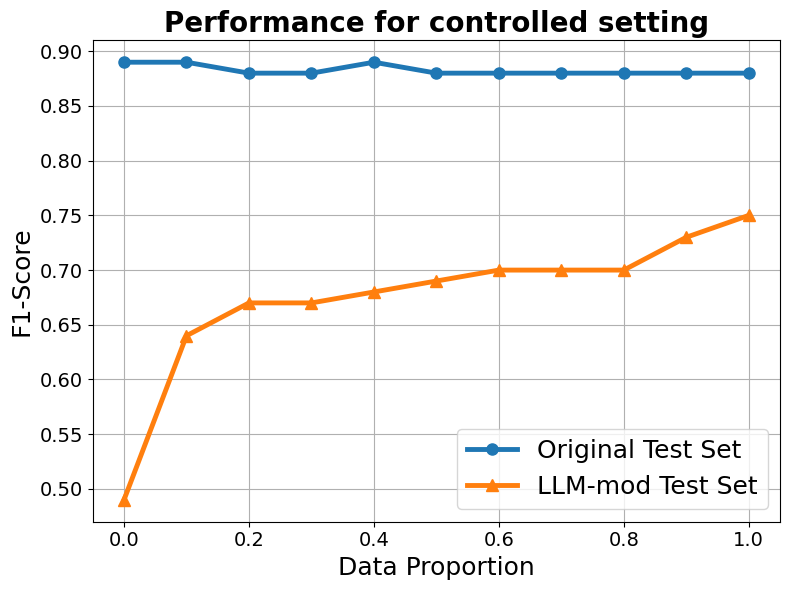}}

     \caption{\textit{Original} and \textit{LLM-modified} test set performance for two model adaptation strategies: (a) Uncontrolled approach uses only LN and LLM-modified PS articles for training; (b) Controlled approach incorporates a subset of original PS articles as a replay buffer to mitigate forgetting. }
    \label{fig:cl_compare}

\end{figure}
\paragraph{Results} 

Figure~\ref{fig:cl_compare} presents the comparative performance of the \textit{controlled} and \textit{uncontrolled} continual learning settings under adversarial drift. To observe the performance, we build two sets: the \textit{original} test set, where both the LN and PS are human-generated, and the \textit{LLM-mod} test set, where PS is LLM-modified. 

In the absence of any LLM-modified PS samples during training, the F1-score on the adversarial test set drops substantially to 49\%, highlighting the model’s vulnerability to distributional shift. As LLM-modified data is incrementally introduced into training, both settings exhibit significant performance gains. Specifically, incorporating 50\% of the LLM-modified training samples boosts the F1-score by 22\% in the uncontrolled setting and by 20\% in the controlled setting. When the full set of LLM-modified samples (100\%) is included, the F1-score reaches 75\% in the uncontrolled setting and 74\% in the controlled setting.

Interestingly, despite the performance improvements on the adversarial test set, the degradation on the original test distribution remains minimal. The uncontrolled setting incurs a modest 3\% decline in F1-score, whereas the controlled setting by leveraging rehearsal via $\mathcal{D}^{(t)}_{\text{PS-base}}$ limits the regression to just 1\%. This suggests that incorporating a partial replay buffer of original PS samples effectively mitigates catastrophic forgetting and helps preserve generalization on the original distribution.

\begin{figure*}[htbp]
    \centering
    \subfigure[]{\includegraphics[width=0.40\textwidth]{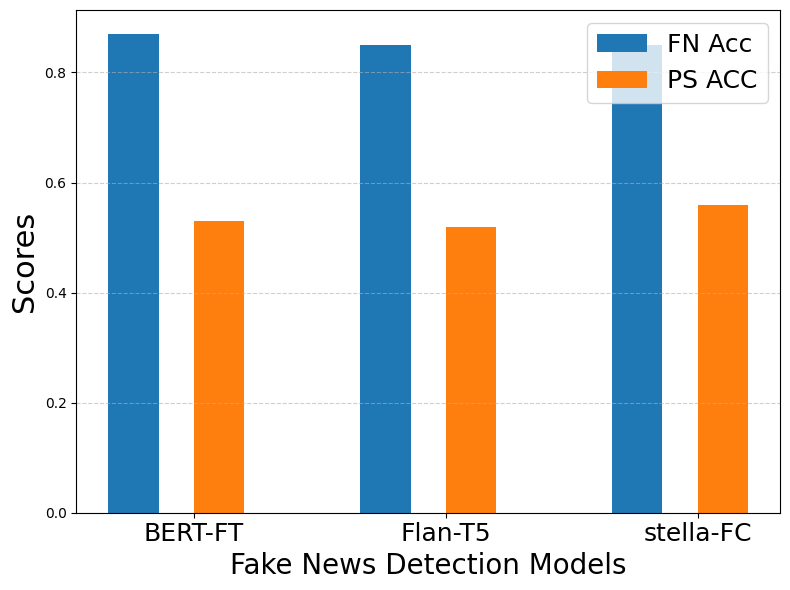}} 
    \subfigure[]{\includegraphics[width=0.40\textwidth]{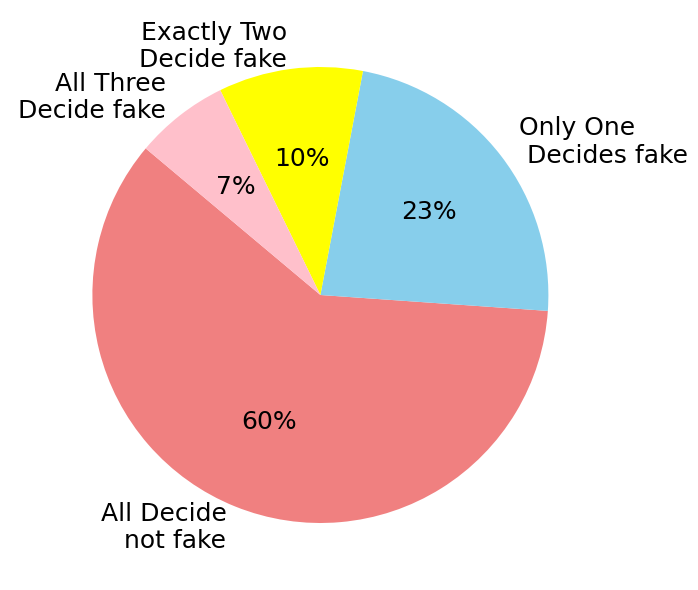}}

     \caption{(a) Comparison of fake news detection models' performance on Pink Slime articles and legitimate news detection. (b) Distribution of each fake news detection model's ``opinion'' about the PS articles being FN. }
    \label{fig:fn-ps}

\end{figure*}

\section{Are Pink Slime articles simply fake news?}

As indicated by Horne et al., PS news articles might not be strictly ``fake news'' \cite{horne2024nela}. While there are evidences that suggest malicious intent underlying pink slime networks, their content often comprises a mix of syndicated authentic news, automated articles sourced from legitimate datasets, and partisan framing, which makes it hard to claim as fake news \cite{horne2024nela}. To rigorously investigate this hypothesis, we employ three state-of-the-art (SOTA) transformer-based fake news detection models: a fully fine-tuned BERT, Flan-T5, and Stella\_en\_v5 embeddings coupled with a three-layer fully connected classification head.

Our findings reveal that for 60\% of Pink Slime articles, none of the models classifies them as fake news (Figure \ref{fig:fn-ps}). Conversely, in only 7\% of cases does all three models unanimously identify Pink Slime articles as fake news. When comparing Pink Slime articles to local news, the classification accuracy across the models was relatively low, achieving 53\%, 51\%, and 56\% for BERT, Flan-T5, and Stella\_en\_v5, respectively. These results suggest that, based on textual characteristics, categorizing Pink Slime articles as fake news may not be appropriate or accurate.

\section{Limitations}
We do not conduct any fact-checking on the datasets used in this study and relied on the provided labels as-is. Consequently, we cannot make any claims about the factual accuracy of the Pink Slime news articles analyzed. While the full NELA-PS dataset contains over 8 million entries, our analysis was limited to a subset of 40,000 articles.

\section{Conclusion}

In this work, we present the first comprehensive linguistic analysis of Pink Slime journalism, revealing consistent syntactic simplicity, limited lexical richness, and templated writing patterns that set it apart from legitimate local news. These insights enabled the development of robust detection models leveraging both handcrafted features and transformer-based fine-tuning approaches. Our study also uncovers an imminent threat: the ability of LLMs to adversarially rephrase the PS articles, significantly degrading the performance of detection systems. To address this, we introduced a continual learning framework that incrementally adapts to LLM-induced distributional shifts while preserving performance on original data. This approach mitigates catastrophic forgetting and improves detection under adversarial conditions by up to 27\%.

To conclude, our work underscores the evolving nature of automated misinformation and provide a scalable framework for resilient Pink Slime detection.



\bibliographystyle{acl_natbib}
\bibliography{ranlp2023}


\end{document}